\documentclass[11pt]{article}
\usepackage[margin=1in]{geometry}
\usepackage{times}
\usepackage[round,authoryear]{natbib}

\usepackage{amsmath,amsfonts,bm}

\def\eqref#1{equation~\ref{#1}}

\def\1{\bm{1}}

\DeclareMathAlphabet{\mathsfit}{\encodingdefault}{\sfdefault}{m}{sl}
\SetMathAlphabet{\mathsfit}{bold}{\encodingdefault}{\sfdefault}{bx}{n}

\usepackage{amsmath,amssymb}
\usepackage{booktabs}
\usepackage{graphicx}
\usepackage{array}
\usepackage{multirow}
\usepackage{xcolor}
\usepackage{hyperref}
\usepackage{microtype}
\usepackage{float}
\usepackage{placeins}

\hypersetup{
  colorlinks=true,
  linkcolor=blue,
  citecolor=blue,
  urlcolor=blue,
  hypertexnames=false,
  pdftitle={Parallel Causal Associative Fields: Gated Sparse Memory for Long-Context Language Modeling},
  pdfauthor={Ahmed},
  pdfsubject={Parallel associative-memory language modeling}
}

\title{Parallel Causal Associative Fields:\\
Gated Sparse Memory for Long-Context Language Modeling}

\author{Ahmed\\
Independent Researcher\\
\texttt{filliones@gmail.com}\\
\url{https://github.com/ahmed123hds/PCAF}
}
\date{June 2026}

\newcommand{\tableformat}{\footnotesize\setlength{\tabcolsep}{4pt}\renewcommand{\arraystretch}{1.08}}
\newcommand{\widedataformat}{\scriptsize\setlength{\tabcolsep}{3pt}\renewcommand{\arraystretch}{1.04}}

\begin{document}
\raggedbottom
\maketitle

\begin{abstract}
Transformers achieve strong language modeling performance by providing direct
token-to-token communication paths, but causal self-attention scales
quadratically with context length. Recurrent and state-space models reduce this
cost, yet compress history into sequentially updated fixed-size states. This
paper studies a third primitive: a parallel content-addressed memory over causal
successor records. The proposed Parallel Causal Associative Field (PCAF) writes
local records from a context window into hash buckets, retrieves a bounded
candidate set for the current query, forms a sparse cache distribution over
successor tokens, and mixes that cache with a parametric local language model
through a learned gate. The resulting model maintains sparse long-context access
while avoiding a single fixed recurrent state bottleneck.
We evaluate PCAF under full autoregressive pretraining on WikiText-103 and
PG-19 using a distributed Google Cloud TPU v4-32 pod. At 303M parameters and
context length $T{=}2048$, PCAF-semantic reaches 36.31 perplexity on
WikiText-103 and 52.45 perplexity on PG-19, compared with 47.49 and 53.84 for
a matched dense Transformer. PCAF-semantic simultaneously processes
0.61--0.62M tokens/s across the TPU pod, versus 0.43M tokens/s for dense and
local attention baselines. Supporting 41M-parameter multi-seed sweeps and
single-GPU component ablations show that the associative cache, retrieval
capacity, and learned gate materially affect the speed--quality trade-off.
\end{abstract}

\section{Introduction}

Modern sequence models face a fundamental tension between memory access breadth
and computational cost. Self-attention grants each token a direct path to every
preceding token, a property that underlies the strong empirical performance of
Transformer-based language models
\citep{vaswani2017attention,bahdanau2015attention}. This direct access is
expensive: a causal attention layer over a length-$T$ sequence requires
$O(T^2 d)$ operations per layer, limiting practical context lengths under
memory and time constraints. Numerous strategies have been proposed to reduce
this cost. Sparse attention methods such as Longformer, BigBird, and the Sparse
Transformer restrict the attention graph to structured subsets of token pairs
\citep{beltagy2020longformer,zaheer2020bigbird,child2019sparsetransformer}.
Linear attention approximates the softmax kernel to achieve $O(T)$ per-layer
complexity \citep{katharopoulos2020linearattn}. Recurrent architectures based
on structured state spaces, including S4 and Mamba, achieve linear-time
sequence processing through selective state updates
\citep{gu2021s4,gu2023mamba}. Each of these approaches makes a distinct
trade-off: sparse attention limits the communication graph, linear attention
approximates the full attention matrix, and recurrent state-space models must
propagate long-range information through compressed fixed-dimensional states.

The motivation of this work is to disentangle two roles that are typically
conflated within the attention mechanism: \emph{memory addressing} and
\emph{value resolution}. Rather than constructing dense token-to-token
interactions, PCAF writes a sparse associative memory of causal records and
performs a bounded read from that memory. This read behaves analogously to a
neural cache \citep{grave2016continuous,khandelwal2019knn}, but is trained
jointly with a local parametric model and implemented as a parallel hash-bucket
primitive rather than an external nearest-neighbor index. The design draws
inspiration from memory-augmented neural networks
\citep{weston2015memory,sukhbaatar2015end2end} and from retrieval-augmented
language models \citep{borgeaud2022retro,izacard2022atlas}, but operates
\emph{within} a single context window without an external corpus.

\paragraph{Biological Inspiration and Design Rationale.}
Complementary learning systems theory distinguishes two forms of memory
computation \citep{mcclelland1995complementary,kumaran2016learning}. The
hippocampal system is associated with rapid binding of individual episodes and
pattern completion from partial cues. Neocortical learning proceeds more
gradually and extracts statistical structure shared across repeated
experiences. This distinction motivates an architecture in which instance-level
retrieval and parametric generalization are implemented by separate,
cooperating components.

PCAF adopts this functional separation. The associative cache stores causal
successor records and retrieves a bounded set of records whose addresses match
the current context. It therefore provides a rapid, instance-specific memory
path. The local parametric network learns reusable syntactic and semantic
regularities and can produce a prediction when no reliable cache record is
available. The learned gate combines the two predictive distributions at each
position, allowing the model to vary its reliance on retrieved evidence and
parametric computation.

This correspondence is a computational analogy, not a claim that PCAF models
the biological mechanisms of the hippocampus or neocortex. Its role is to state
the design objective clearly: retain direct access to selected past events
without requiring dense comparison against every preceding token, while
preserving a parametric path for generalization beyond stored contexts.

The central empirical question is practical: can a simple associative memory
recover useful long-context signal without paying the cost of dense attention?
We evaluate this question primarily with full autoregressive pretraining on
WikiText-103 and PG-19, then use smaller controlled experiments to isolate
the effects of routing, cache capacity, and the learned mixture gate.

\paragraph{Contributions.}
This paper makes four concrete contributions:
\begin{enumerate}
  \item \textbf{A new causal sequence-modeling primitive.}
  We introduce the Parallel Causal Associative Field (PCAF), which replaces
  dense token-to-token attention with a parallel associative-memory read over
  causal successor records. Unlike recurrent and state-space models, PCAF does
  not compress the entire prefix into a single fixed-size state; unlike dense
  attention, it does not score every past token as a value source. Instead, it
  decouples \emph{addressing} from \emph{value resolution}: a discrete or
  learned route selects a bounded candidate set, and a continuous key--query
  score resolves only those candidates.

  \item \textbf{A gated sparse cache architecture for language modeling.}
  We instantiate PCAF as a language model with two parallel paths: a local
  causal parametric path and a sparse successor-token cache. The final
  distribution is a learned mixture of parametric prediction and retrieved
  cache probability at every token position. This gate allows the model to use
  associative recall when the current context has useful predecessor records,
  while falling back to parametric composition when retrieval is unreliable.

  \item \textbf{Semantic and discrete routing mechanisms with bounded
  retrieval.}
  We study both exact context-hash routing and learned semantic routing. The
  context route gives fast, high-precision discrete cache hits; the semantic
  route addresses the surface-form limitation of exact token hashing by allowing
  hidden-state-similar contexts to retrieve one another even when their surface
  forms differ. In both cases, retrieval is bounded by a small top-$K$ candidate
  set, yielding sparse memory access rather than dense attention.

  \item \textbf{Full autoregressive evidence at matched scale.}
  We evaluate PCAF under full autoregressive pretraining, not only next-token
  diagnostics. At approximately 303M parameters on WikiText-103 and PG-19,
  PCAF achieves lower perplexity than matched dense Transformer and local
  attention baselines while processing more tokens per second on a TPU v4-32
  pod. Additional 41M-parameter multi-seed experiments and RTX 3060 profiling
  isolate the effects of cache routing, gating, and hardware efficiency.
\end{enumerate}

\section{Related Work}

\paragraph{Self-attention and long-context Transformers.}
The Transformer architecture replaces recurrence with self-attention, enabling
parallel communication between all token pairs
\citep{vaswani2017attention}. For long contexts, this dense communication is
prohibitively expensive. Transformer-XL extends context through segment-level
recurrence and relative-position encodings \citep{dai2019transformerxl}.
Longformer combines a sliding-window local attention with task-specific global
tokens \citep{beltagy2020longformer}; BigBird adds random attention edges and
studies the theoretical expressivity of sparse patterns
\citep{zaheer2020bigbird}. The Sparse Transformer \citep{child2019sparsetransformer}
introduces strided and fixed factorizations of the attention matrix. Positional
encoding improvements, including ALiBi \citep{press2021alibi} and RoPE
\citep{su2021roformer}, partially mitigate length-generalization failures but do
not change the asymptotic complexity of attention. PCAF does not attend over a
fixed sparse graph at all; instead it writes and reads causal successor records
from a content-addressed hash memory.

\paragraph{Linear and subquadratic attention.}
Linear attention methods reformulate the softmax attention kernel so that key
and value products can be accumulated left-to-right, reducing per-layer
complexity from $O(T^2 d)$ to $O(T d^2)$
\citep{katharopoulos2020linearattn}. While this achieves linear time, the
approximation can degrade model quality relative to softmax attention at
comparable scale. PCAF sidesteps the approximation altogether by reading only a
small bucket of $K$ candidates, trading recall for precision in a way that is
empirically validated through controlled ablations.

\paragraph{State-space and recurrent models.}
Structured state-space models (S4) achieve efficient long-range sequence
modeling through convolutions with HiPPO-initialized kernels
\citep{gu2021s4}. Mamba extends this with input-dependent selective state
updates and hardware-aware parallel scan kernels, matching Transformer quality
at linear cost \citep{gu2023mamba}. Mamba-2 \citep{dao2024mamba2} establishes
a theoretical duality between state-space models and structured attention.
RWKV \citep{peng2023rwkv} adopts a linear recurrence with token-mixing
analogous to time-decay attention. RetNet \citep{sun2023retnet} introduces
retention, a decay-weighted recurrence that admits both parallel and sequential
formulations. These models represent the history in a fixed-size state vector,
which may discard fine-grained token-level patterns. PCAF instead stores
discrete successor records and reads a bounded candidate set by address,
avoiding the fixed-state bottleneck at the cost of per-window memory
allocation.

\paragraph{Sparse random access for recurrent models.}
RAMba augments Mamba with hardware-aligned hierarchical sparse attention to
restore random access to long histories \citep{hu2025ramba}. It partitions the
sequence into chunks, selects relevant chunks using token-level relevance
signals, and hierarchically aggregates their contents. This approach and PCAF
share the objective of avoiding exclusive dependence on a fixed recurrent
state. Their memory mechanisms differ: RAMba adds learned sparse attention over
selected chunks, whereas PCAF constructs causal successor records and performs
a bounded content-addressed read whose retrieved token distribution is combined
with a local parametric model.

\paragraph{Cache and retrieval language models.}
Cache language models exploit the burstiness of natural language by boosting
the probability of recently observed words. Continuous cache models store past
hidden states and access them through dot-product similarity
\citep{grave2016continuous}. $k$NN-LMs interpolate a base neural language model
with a nearest-neighbor distribution over an external datastore
\citep{khandelwal2019knn}. RETRO \citep{borgeaud2022retro} retrieves
text chunks from a large offline corpus and attends to them with
cross-attention. ATLAS \citep{izacard2022atlas} integrates retrieval into
few-shot learning through joint encoder-retriever training. PCAF is closest in
spirit to cache language models. The architectural difference is that the cache
is constructed \emph{within} the current sequence window during the forward
pass, uses a parallel hash-bucket candidate selection implemented as a custom
Triton kernel, and is mixed with the parametric path through a learned gate
trained end-to-end without any external index or offline retrieval step.

\paragraph{Memory-augmented neural networks.}
Memory Networks \citep{weston2015memory} and End-to-End Memory Networks
\citep{sukhbaatar2015end2end} demonstrated that explicit addressable memory
modules can improve reasoning over longer contexts. Neural Turing Machines and
Differentiable Neural Computers extend this idea to differentiable read/write
operations \citep{graves2014neural}. PCAF borrows the spirit of content-based
addressing from this line of work but replaces differentiable addressing with a
discrete hash-bucket scheme that is more amenable to GPU-parallel
implementation and does not require end-to-end gradient flow through the memory
indices.

\section{Method}

\begin{figure}[H]
  \centering
  \includegraphics[width=0.66\linewidth]{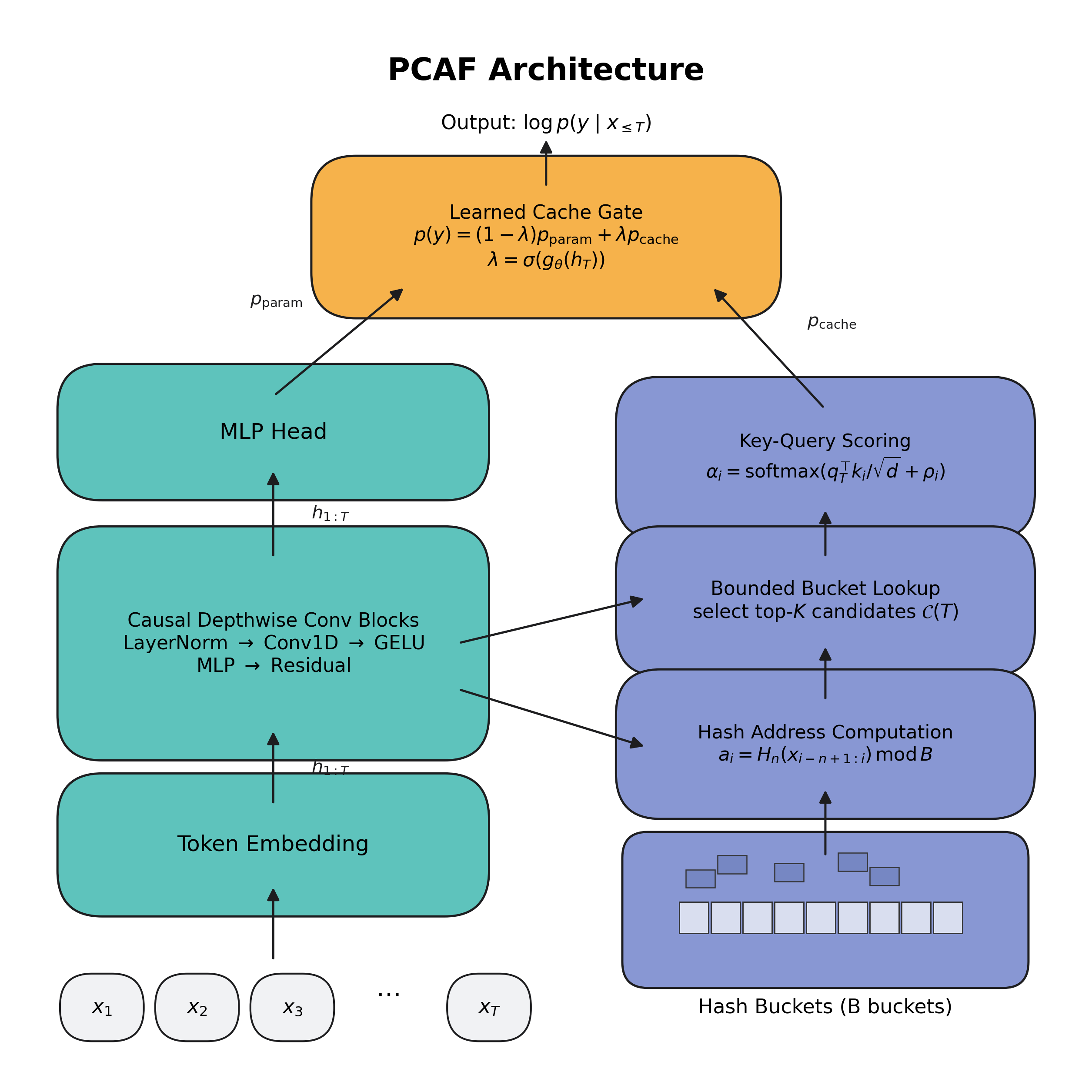}
\caption{PCAF architecture. A causal local network produces hidden states and a
parametric token distribution. In parallel, an address function routes causal
successor records into hash buckets. At most $K$ retrieved records are scored
against the current query to form $p_{\mathrm{cache}}$, and a learned gate
combines it with $p_{\mathrm{param}}$. Full autoregressive training applies this
computation at every causal position.}
  \label{fig:architecture}
\end{figure}

Figure~\ref{fig:architecture} illustrates the overall architecture of the proposed Parallel Causal Associative Field (PCAF) model. The architecture consists of a local parametric path running in parallel with a content-addressed associative memory module. Under this scheme, memory addressing and value resolution are decoupled, allowing fast subquadratic sequence processing with bounded candidate lookups.

\subsection{Problem Setup}

Given a token sequence $x_{1:T}$, the full autoregressive objective predicts
each token from its causal prefix:
\begin{equation}
  \mathcal{L}_{\mathrm{AR}} =
  -\frac{1}{T}\sum_{t=1}^{T}\log p(x_t \mid x_{<t}).
\end{equation}
Our headline experiments use this objective. We also report controlled
next-token diagnostic runs, where only $x_{T+1}$ is predicted from
$x_{1:T}$, to isolate routing and cache behavior at long context lengths.

\subsection{Local Parametric Path}

PCAF-context begins with token embeddings $e_t \in \mathbb{R}^d$ and passes
them through a small stack of causal depthwise convolutional blocks:
\begin{equation}
  h_{1:T} = f_{\theta}(e_{1:T}),
\end{equation}
where each block applies layer normalization, causal depthwise convolution
(kernel size 5), a pointwise two-layer MLP with GELU activation, dropout, and a
residual connection. Causality is enforced by left-padding the convolution
input by $(\text{kernel size} - 1)$ positions before applying the depthwise
filter. The final hidden state $h_T$ produces the parametric language-model
logits:
\begin{equation}
  p_{\mathrm{param}}(y \mid x_{\leq T}) =
  \mathrm{softmax}(W_o \, \phi(h_T)),
\end{equation}
where $\phi$ is a two-layer MLP head with layer normalization and GELU.
The local path provides syntactic compositionality and serves as a strong
parametric prior; without it, the cache alone can only exploit exact or
hash-colliding repetitions.

\subsection{Associative Successor Memory}

For each position $i < T$, PCAF constructs a causal successor record:
\begin{equation}
  \mathrm{rec}_i = (a_i,\; k_i,\; v_i),
\end{equation}
where $a_i \in \{0,\ldots,B-1\}$ is a discrete bucket address,
$k_i \in \mathbb{R}^d$ is a continuous key, and $v_i = x_{i+1}$ is the integer
successor token. The address is computed from a causal $n$-gram hash:
\begin{equation}
  a_i = H_n(x_{i-n+1:i}) \bmod B,
\end{equation}
where $H_n$ is a polynomial rolling hash over $n$ tokens with zero-padding at
the left boundary. In the best-performing token-hash configuration, $n=1$. The
continuous keys are $\ell_2$-normalized linear projections of the local hidden
states:
\begin{equation}
  k_i = \mathrm{norm}(W_k h_i), \qquad q_T = \mathrm{norm}(W_q h_T).
\end{equation}

Exact token hashes do not capture semantic similarity. For example,
paraphrastic contexts such as ``red car'' and ``scarlet automobile'' receive
unrelated addresses even if their hidden states are nearby. We therefore also
implement a learned semantic route. A router maps hidden states to distributions
over semantic buckets,
\begin{equation}
  \tilde{r}_i = \mathrm{softmax}(W_s h_i / \tau), \qquad
  \tilde{r}_T = \mathrm{softmax}(W_s h_T / \tau),
\end{equation}
and scores candidate records by semantic overlap
$m_i = \tilde{r}_T^\top \tilde{r}_i$. The model supports semantic candidates
alone or a hybrid union of exact hash candidates and semantic candidates. In the
hybrid route, exact matching preserves high-precision cache hits while semantic
routing allows hidden-state similar contexts to retrieve each other even when
their surface forms differ.

\noindent\begin{minipage}{\linewidth}
The forward pass proceeds as follows:
\begin{enumerate}
  \item $e_{1:T} \leftarrow \texttt{Embedding}(x_{1:T})$
  \item $h_{1:T} \leftarrow f_\theta(e_{1:T})$ \hfill\textit{(local causal conv blocks)}
  \item $p_{\mathrm{param}} \leftarrow \mathrm{softmax}(W_o\,\phi(h_T))$
  \item $a_{1:T-1} \leftarrow H_n(x_{1:T-1}) \bmod B$; $\quad a_T \leftarrow H_n(x_{1:T}) \bmod B$
  \item $k_{1:T-1} \leftarrow \mathrm{norm}(W_k h_{1:T-1})$; $\quad q_T \leftarrow \mathrm{norm}(W_q h_T)$
  \item $\mathcal{C}(T) \leftarrow \texttt{TritonBucketLookup}(a_{1:T-1},\,a_T,\,K,\,B)$
  \item $\alpha_i \leftarrow \mathrm{softmax}_i\!\bigl(q_T^\top k_i / {\sqrt{d}} + \rho\,i/(T{-}1)\bigr)$, $\;i\in\mathcal{C}(T)$
  \item $p_{\mathrm{cache}}(y) \leftarrow \sum_{i \in \mathcal{C}(T)} \alpha_i\,\mathbf{1}[x_{i+1}=y]$
  \item $\lambda_T \leftarrow \sigma(g_\theta(h_T))\cdot\mathbf{1}[\lvert\mathcal{C}(T)\rvert>0]$
  \item \textbf{return} $\log\!\bigl[(1-\lambda_T)\,p_{\mathrm{param}} + \lambda_T\,p_{\mathrm{cache}}\bigr]$
\end{enumerate}
\end{minipage}

The query address $a_T$ selects one hash bucket. A custom Triton kernel builds
fixed-size bucket index arrays in parallel over all sequence positions and
returns the indices of at most $K$ candidate records from the selected bucket.
The learned scalar recency bias $\rho$ in step 7 above biases retrieval toward
more recent records within the same bucket. Normalized cache weights are:
\begin{equation}
  \alpha_i =
  \frac{\exp(s_i)}{\sum_{j \in \mathcal{C}(T)} \exp(s_j)},
  \qquad i \in \mathcal{C}(T),
\end{equation}
where $s_i = q_T^\top k_i / \sqrt{d} + \rho\, i/(T-1)$.
These weights define a sparse distribution over successor tokens:
\begin{equation}
  p_{\mathrm{cache}}(y \mid x_{\leq T})
  = \sum_{i \in \mathcal{C}(T)} \alpha_i \;\mathbf{1}[v_i = y].
  \label{eq:cache-dist}
\end{equation}

\subsection{Learned Cache Gate}

The final predictive distribution is a learned convex combination of the
parametric path and the cache:
\begin{align}
  \lambda_T &= \sigma\bigl(g_{\theta}(h_T)\bigr), \\
  p(y \mid x_{\leq T}) &=
  (1 - \lambda_T)\,p_{\mathrm{param}}(y \mid x_{\leq T})
  + \lambda_T\,p_{\mathrm{cache}}(y \mid x_{\leq T}),
  \label{eq:mixture}
\end{align}
where $g_\theta$ is a two-layer MLP with sigmoid output. When the selected
bucket is empty, the gate is masked to zero and the model falls back entirely
to the parametric path. For a single queried position, the loss is next-token
cross-entropy:
\begin{equation}
  \mathcal{L} = -\log p(x_{T+1} \mid x_{\leq T}).
\end{equation}
The full autoregressive experiments apply this prediction rule at every
position and average the losses as in Section~3.1; the diagnostic experiments
apply it only at the final position of each sampled context window.
For numerical stability, the mixture is computed in log-space via
\texttt{logaddexp}, preventing underflow in the sparse cache term.

\subsection{Computational Cost}

The associative memory read cost per sample is $O(T + Kd)$ with $K \ll T$.
Dense causal attention costs $O(T^2 d)$ per layer. Sliding-window attention
reduces this to $O(Twd)$ per layer for window size $w$, but applies attention
in every layer. PCAF replaces the multi-layer attention stack with a
linear-cost local convolutional path plus one sparse memory read, yielding the throughput
advantages reported in Table~\ref{tab:submission_303m_results} and the
diagnostic scaling results in Tables~\ref{tab:tpu_baselines}--\ref{tab:tpu_ablations}.

\section{Experiments}

\subsection{Datasets}

Experiments are conducted on three language-modeling benchmarks.
\textbf{WikiText-103} \citep{merity2016pointer} is the primary large-scale
benchmark for full autoregressive pretraining; it contains approximately
103~million training tokens from Wikipedia articles. \textbf{PG-19} is used as
a second long-form benchmark to test whether the same speed--quality trade-off
holds on book-length text. \textbf{WikiText-2} \citep{merity2016pointer} is
reserved for controlled single-GPU ablations that isolate the cache and gate
components under a consumer-GPU memory budget.

\subsection{Experimental Setup}

\paragraph{Primary hardware: Google Cloud TPU v4-32 pod.}
All headline full autoregressive experiments are conducted on a
\textbf{Google Cloud TPU v4-32 cluster} (32 TPU v4 cores, 1,024~GB HBM in
total). Models are implemented in JAX and parallelized with \texttt{jax.pmap}.
The TPU implementation uses XLA tensor primitives for routing and candidate
selection rather than the CUDA-specific Triton bucket kernel. Throughput is
reported across the full pod in tokens per second.

\paragraph{Secondary hardware: single NVIDIA RTX 3060 GPU.}
Controlled ablation experiments on WikiText-2 are conducted on a single
\textbf{NVIDIA RTX 3060} GPU with 12\,GB VRAM and 64\,GB system RAM, using a
PyTorch implementation with a custom Triton bucket-lookup kernel. This setting
is used only to isolate the contribution of individual PCAF components.

\paragraph{Baselines and budgets.}
The main 303M-parameter comparison uses a dense Transformer, a 128-token local
Transformer, a local-convolution-only model, PCAF-context, and PCAF-semantic at
sequence lengths $T{=}1024$ and $T{=}2048$. The 41M-parameter experiments add
linear attention, global-local attention, bucket-count ablations, top-$K$
ablations, and sequence length scaling. Unless otherwise stated, all compared
models in a table share the same corpus, tokenizer, sequence length, training
budget, and validation selection rule. We report both final validation
perplexity and best validation perplexity; the latter is the model-selection
metric used when a run begins to overfit late in training.

\subsection{Main Results: Full Autoregressive Pretraining at 303M Parameters}
\label{sec:full-ar-303m}

Table~\ref{tab:submission_303m_results} is the main result table of the paper.
It reports full autoregressive language modeling at 303M parameters on
WikiText-103 and PG-19 at sequence lengths $T = 1024$ and $T = 2048$.
Across these configurations, PCAF gives lower perplexity and higher throughput
than the matched dense and local attention baselines.

Memory use partly explains the throughput difference. At sequence length
$T = 2048$, the dense and local Transformer implementations run out of memory
when the global batch size exceeds 32 on the TPU v4-32. The PCAF and local
convolutional implementations fit a global batch size of 256 because they do
not materialize a quadratic attention matrix.

On WikiText-103 at $T = 1024$, PCAF-semantic obtains a best perplexity of
\textbf{33.30} PPL (final 33.43 PPL), which is 16.22 points below the dense
Transformer and 8.38 points below the local convolutional baseline. Its measured
throughput is 0.89M tokens/s at global batch size 256, compared with 0.44M
tokens/s for the dense Transformer at global batch size 32. At $T = 2048$ and
equal global batch size 32, PCAF-semantic obtains \textbf{36.31} PPL and 0.61M
tokens/s, compared with 47.49 PPL and 0.43M tokens/s for the dense Transformer.
At global batch size 256, which does not fit for the attention baselines in this
implementation, PCAF-semantic processes 0.89M tokens/s.

On PG-19 at $T = 1024$, PCAF-semantic obtains a best perplexity of
\textbf{50.94} PPL (final 59.85 PPL), 5.36 points below the dense Transformer
and 12.85 points below the local convolutional baseline. At $T = 2048$ and
global batch size 32, PCAF-semantic obtains \textbf{52.45} PPL at 0.62M
tokens/s; the dense Transformer obtains 53.84 PPL at 0.43M tokens/s, and the
local Transformer obtains 54.54 PPL at 0.43M tokens/s. These measurements show
the quality and throughput differences separately from the additional
throughput available to PCAF at larger batch sizes.

\begin{table}[H]
  \centering
  \caption{Large-scale full autoregressive JAX results on a Google Cloud TPU v4-32 cluster. All models are trained at sequence lengths $T=1024$ and $T=2048$; throughput is measured across the full TPU pod in millions of tokens per second (M tok/s). The number in parentheses denotes the global batch size. Dense and local attention baselines trigger out-of-memory (OOM) errors at batch size 256 for sequence length $T=2048$, limiting them to a maximum batch size of 32.}
  \label{tab:submission_303m_results}
  \begingroup
  \tableformat
  \resizebox{\linewidth}{!}{
  \begin{tabular}{@{}lllrrrrrr@{}}
    \toprule
    Dataset & Model & Seq. ($T$) & Seeds & Params & Best PPL $\downarrow$ & Final PPL $\downarrow$ & Acc. @ Best $\uparrow$ & Throughput (Batch Size) \\
    \midrule
    PG-19 & Dense Transformer & 1024 & 1 & 301.85M & 56.30 & 56.30 & 29.17\% & 0.44M (32) \\
          &                   & 2048 & 1 & 302.77M & 53.84 & 53.84 & 29.39\% & 0.43M (32) \\
          \cmidrule{2-9}
          & Local Transformer & 1024 & 1 & 301.85M & 57.32 & 57.32 & 29.06\% & 0.44M (32) \\
          &                   & 2048 & 1 & 302.77M & 54.54 & 54.54 & 29.47\% & 0.43M (32) \\
          \cmidrule{2-9}
          & Local conv        & 1024 & 1 & 303.27M & 63.79 & 76.67 & 27.03\% & 1.20M (256) \\
          &                   & 2048 & 1 & 303.27M & 61.99 & 61.99 & 28.44\% & 0.78M (32) / 1.20M (256) \\
          \cmidrule{2-9}
          & PCAF-context      & 1024 & 1 & 303.27M & 59.31 & 71.42 & 26.64\% & 0.96M (256) \\
          &                   & 2048 & 1 & 303.27M & 57.00 & 57.00 & 27.91\% & 0.65M (32) / 0.96M (256) \\
          \cmidrule{2-9}
          & PCAF-semantic     & 1024 & 1 & 303.27M & \textbf{50.94} & 59.85 & 26.13\% & 0.89M (256) \\
          &                   & 2048 & 1 & 303.27M & \textbf{52.45} & 52.45 & 27.69\% & 0.62M (32) / 0.89M (256) \\
    \midrule
    WikiText-103 & Dense Transformer & 1024 & 1 & 301.85M & 49.52 & 49.52 & 32.19\% & 0.44M (32) \\
                 &                   & 2048 & 1 & 302.77M & 47.49 & 47.49 & 32.57\% & 0.43M (32) \\
                 \cmidrule{2-9}
                 & Local Transformer & 1024 & 1 & 301.85M & 51.00 & 51.00 & 32.06\% & 0.44M (32) \\
                 &                   & 2048 & 1 & 302.77M & 44.42 & 44.42 & 33.25\% & 0.43M (32) \\
                 \cmidrule{2-9}
                 & Local conv        & 1024 & 1 & 303.27M & 41.68 & 41.79 & 34.29\% & 1.20M (256) \\
                 &                   & 2048 & 1 & 303.27M & 42.44 & 42.44 & 34.03\% & 0.78M (32) / 1.20M (256) \\
                 \cmidrule{2-9}
                 & PCAF-context      & 1024 & 1 & 303.27M & 38.44 & 38.62 & 34.07\% & 0.96M (256) \\
                 &                   & 2048 & 1 & 303.27M & 37.87 & 37.87 & 33.99\% & 0.65M (32) / 0.96M (256) \\
                 \cmidrule{2-9}
                 & PCAF-semantic     & 1024 & 1 & 303.27M & \textbf{33.30} & 33.43 & 33.65\% & 0.89M (256) \\
                 &                   & 2048 & 1 & 303.27M & \textbf{36.31} & 36.31 & 33.68\% & 0.61M (32) / 0.89M (256) \\
    \bottomrule
  \end{tabular}
  }
  \endgroup
\end{table}

\begin{figure}[H]
  \centering
  \includegraphics[width=0.78\linewidth]{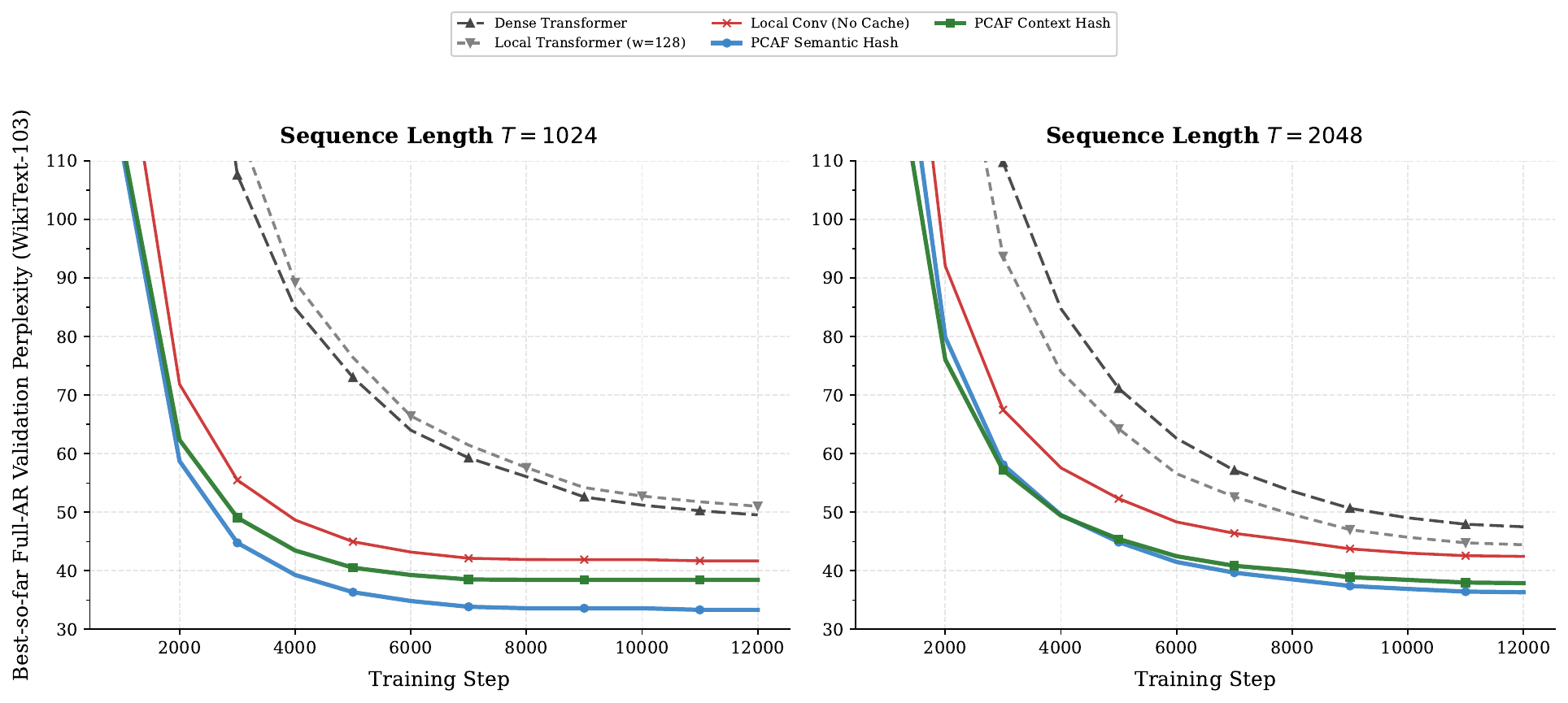}
  \caption{Best-so-far full autoregressive validation perplexity on WikiText-103 at sequence lengths $T=1024$ and $T=2048$ on a TPU v4-32 cluster. The vertical axis focuses on the competitive perplexity regime after the initial high-loss transient; Table~\ref{tab:submission_303m_results} reports the exact best and final values.}
  \label{fig:full_ar_curves_wikitext103}
\end{figure}

\begin{figure}[H]
  \centering
  \includegraphics[width=0.78\linewidth]{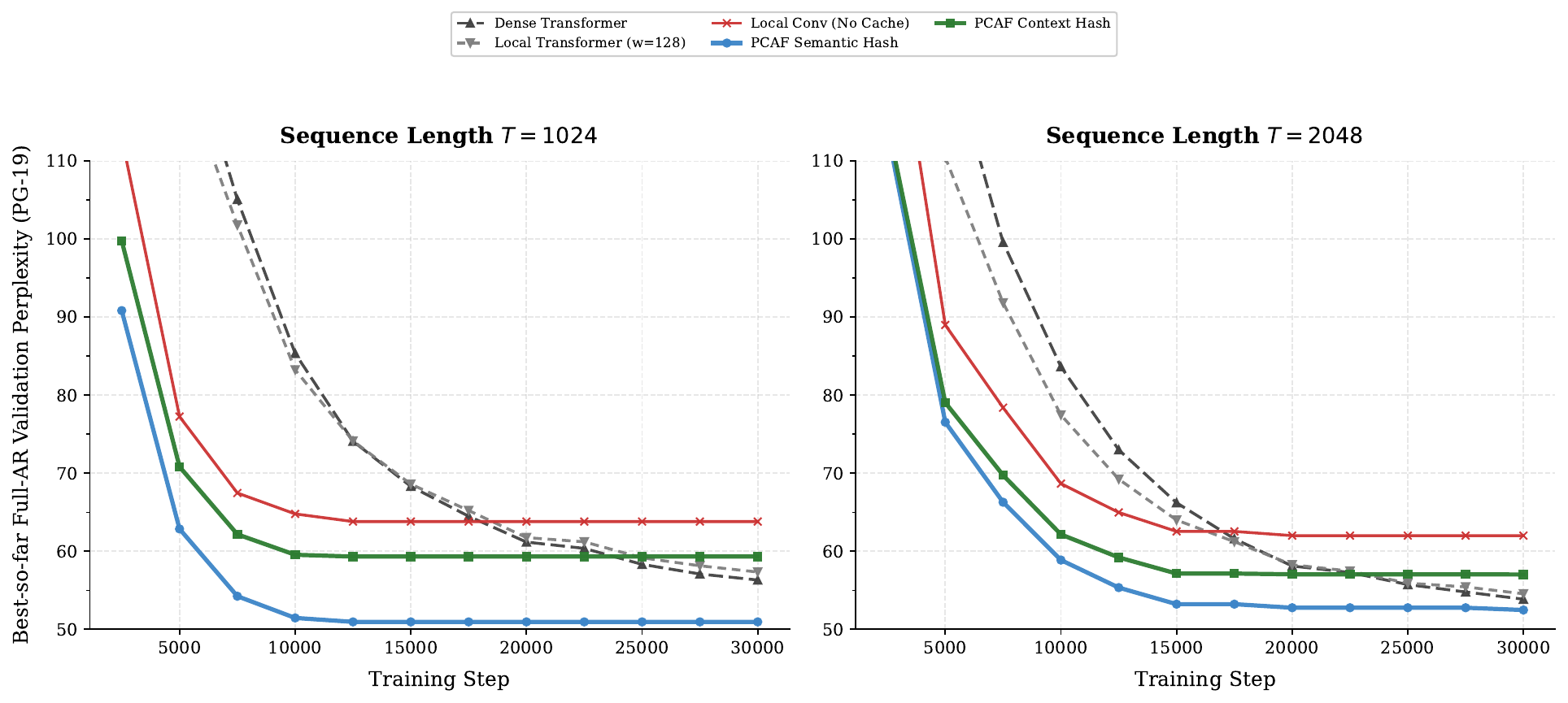}
  \caption{Best-so-far full autoregressive validation perplexity on PG-19 at sequence lengths $T=1024$ and $T=2048$ on a TPU v4-32 cluster. The vertical axis focuses on the competitive perplexity regime, making the lower PCAF-semantic best perplexity visible even when raw validation curves later overfit upward; Table~\ref{tab:submission_303m_results} reports the exact best and final values.}
  \label{fig:full_ar_curves_pg19}
\end{figure}

\subsection{Full Autoregressive Robustness and Sensitivity at 41M Parameters}

Table~\ref{tab:submission_41m_results} provides supporting multi-seed full
autoregressive evidence together with controlled next-token sensitivity
diagnostics. At 41M parameters, PCAF variants retain a large throughput
advantage over attention baselines while remaining stable across seeds. The
next-token bucket and top-$K$ diagnostics show that quality is not driven by a
single fragile memory setting: increasing top-$K$ improves recall, while bucket
counts from 8,192 to 131,072 produce similar perplexities. A separate full-AR
semantic-capacity sweep at $T=2048$ and $K=32$ improves best perplexity from
42.43 to 41.30 as the number of semantic buckets $S$ increases from 128 to
1,024. This gain has a measured systems cost: throughput decreases from 2.87M
to 2.20M tokens/s, while top-1 accuracy remains near 32\%. The $S=512$ setting
is a useful quality--throughput compromise, whereas $S=1024$ gives the best
perplexity. We additionally stress-test long-context full autoregressive
routing at $T=4096$: increasing the retrieval budget from $K=32$ to $K=64$
improves best perplexity from 45.16 to 42.70, while retaining 1.68M tokens/s
on the TPU v4-32 pod.

\begin{table}[H]
  \centering
  \caption{41M-parameter TPU results and sensitivity sweeps. Rows marked ``full AR'' use the full autoregressive loss over all positions. Rows marked ``NT diagnostic'' are next-token-only diagnostic runs from a single sampled context window and are included only to isolate routing, bucket-count, and top-$K$ behavior. Throughput is reported across the TPU v4-32 pod.}
  \label{tab:submission_41m_results}
  \begingroup
  \widedataformat
  \resizebox{\linewidth}{!}{
  \begin{tabular}{@{}llrrrrrr@{}}
    \toprule
    Dataset & Model & Seq. & Seeds & Params & Best PPL & Acc. @ Best & Throughput \\
    \midrule
    PG-19 & Dense Transformer (NT diagnostic) & 2048 & 3 & 41.71M & 238.17 $\pm$ 7.90 & 19.03 $\pm$ 0.44\% & 7.16 $\pm$ 0.00M tok/s \\
    PG-19 & Global-local Transformer (NT diagnostic) & 2048 & 3 & 41.71M & 221.17 $\pm$ 2.48 & 19.46 $\pm$ 0.30\% & 7.15 $\pm$ 0.00M tok/s \\
    PG-19 & Linear attention (NT diagnostic) & 2048 & 3 & 41.71M & 199.62 $\pm$ 1.25 & 20.45 $\pm$ 0.40\% & 11.86 $\pm$ 0.00M tok/s \\
    PG-19 & Local conv (NT diagnostic) & 2048 & 3 & 41.74M & 143.21 $\pm$ 2.41 & 21.94 $\pm$ 0.21\% & 45.66 $\pm$ 0.00M tok/s \\
    PG-19 & PCAF-context (NT diagnostic) & 2048 & 3 & 41.74M & 113.40 $\pm$ 0.86 & 23.49 $\pm$ 0.08\% & 42.81 $\pm$ 0.00M tok/s \\
    PG-19 & PCAF-semantic (NT diagnostic) & 2048 & 3 & 41.74M & 119.85 $\pm$ 0.49 & 22.65 $\pm$ 0.14\% & 41.68 $\pm$ 0.00M tok/s \\
    \midrule
    WikiText-103 & Dense Transformer (NT diagnostic) & 2048 & 3 & 41.71M & 230.40 $\pm$ 7.55 & 22.47 $\pm$ 0.66\% & 7.16 $\pm$ 0.00M tok/s \\
    WikiText-103 & Global-local Transformer (NT diagnostic) & 2048 & 3 & 41.71M & 218.66 $\pm$ 8.15 & 22.87 $\pm$ 1.16\% & 7.15 $\pm$ 0.00M tok/s \\
    WikiText-103 & Linear attention (NT diagnostic) & 2048 & 3 & 41.71M & 168.38 $\pm$ 2.80 & 24.84 $\pm$ 0.71\% & 11.86 $\pm$ 0.00M tok/s \\
    WikiText-103 & Local conv (NT diagnostic) & 2048 & 3 & 41.74M & 92.55 $\pm$ 3.02 & 28.00 $\pm$ 0.29\% & 45.83 $\pm$ 0.00M tok/s \\
    WikiText-103 & PCAF buckets=131072 (NT diagnostic) & 2048 & 1 & 41.74M & 95.54 & 28.03\% & 42.95M tok/s \\
    WikiText-103 & PCAF buckets=32768 (NT diagnostic) & 2048 & 1 & 41.74M & 95.54 & 28.03\% & 42.95M tok/s \\
    WikiText-103 & PCAF buckets=8192 (NT diagnostic) & 2048 & 1 & 41.74M & 95.79 & 27.97\% & 42.95M tok/s \\
    WikiText-103 & PCAF top-k=16 (NT diagnostic) & 2048 & 1 & 41.74M & 95.54 & 28.03\% & 42.95M tok/s \\
    WikiText-103 & PCAF top-k=32 (NT diagnostic) & 2048 & 1 & 41.74M & 93.93 & 28.21\% & 42.81M tok/s \\
    WikiText-103 & PCAF top-k=4 (NT diagnostic) & 2048 & 1 & 41.74M & 99.39 & 28.38\% & 43.09M tok/s \\
    WikiText-103 & PCAF top-k=8 (NT diagnostic) & 2048 & 1 & 41.74M & 97.16 & 28.33\% & 43.00M tok/s \\
    WikiText-103 & PCAF-context (NT diagnostic) & 2048 & 6 & 41.74M & 89.87 $\pm$ 15.28 & 28.61 $\pm$ 0.81\% & 42.96 $\pm$ 0.00M tok/s \\
    WikiText-103 & PCAF-semantic (NT diagnostic) & 2048 & 3 & 41.74M & 79.38 $\pm$ 2.36 & 28.65 $\pm$ 0.44\% & 41.81 $\pm$ 0.00M tok/s \\
    \midrule
    WikiText-103 & PCAF-semantic $S=128$, $K=32$ (full AR) & 2048 & 1 & 41.89M & 42.43 & 32.05\% & 2.87M tok/s \\
    WikiText-103 & PCAF-semantic $S=256$, $K=32$ (full AR) & 2048 & 1 & 41.92M & 42.17 & 32.05\% & 2.70M tok/s \\
    WikiText-103 & PCAF-semantic $S=512$, $K=32$ (full AR) & 2048 & 1 & 41.98M & 41.86 & 31.98\% & 2.55M tok/s \\
    WikiText-103 & PCAF-semantic $S=1024$, $K=32$ (full AR) & 2048 & 1 & 42.09M & \textbf{41.30} & \textbf{32.13\%} & 2.20M tok/s \\
    \midrule
    WikiText-103 & PCAF-semantic top-k=32 (full AR) & 4096 & 1 & 41.92M & 45.16 & 32.11\% & 2.28M tok/s \\
    WikiText-103 & PCAF-semantic top-k=64 (full AR) & 4096 & 1 & 41.92M & 42.70 & 31.95\% & 1.68M tok/s \\
    WikiText-103 & pcaf\_scale\_seq1024 (NT diagnostic) & 1024 & 1 & 41.74M & 96.44 & 27.71\% & 29.19M tok/s \\
    WikiText-103 & pcaf\_scale\_seq2048 (NT diagnostic) & 2048 & 1 & 41.74M & 95.54 & 28.03\% & 42.95M tok/s \\
    WikiText-103 & pcaf\_scale\_seq4096 (NT diagnostic) & 4096 & 1 & 41.74M & 118.45 & 26.41\% & 42.86M tok/s \\
    WikiText-103 & pcaf\_scale\_seq512 (NT diagnostic) & 512 & 1 & 41.74M & 100.41 & 27.77\% & 17.31M tok/s \\
    \bottomrule
  \end{tabular}
  }
  \endgroup
\end{table}
\FloatBarrier

\subsection{Routing and Next-Token Diagnostics on TPU v4-32}

\begin{figure}[H]
  \centering
  \vspace{-0.5em}
  \includegraphics[width=0.82\linewidth]{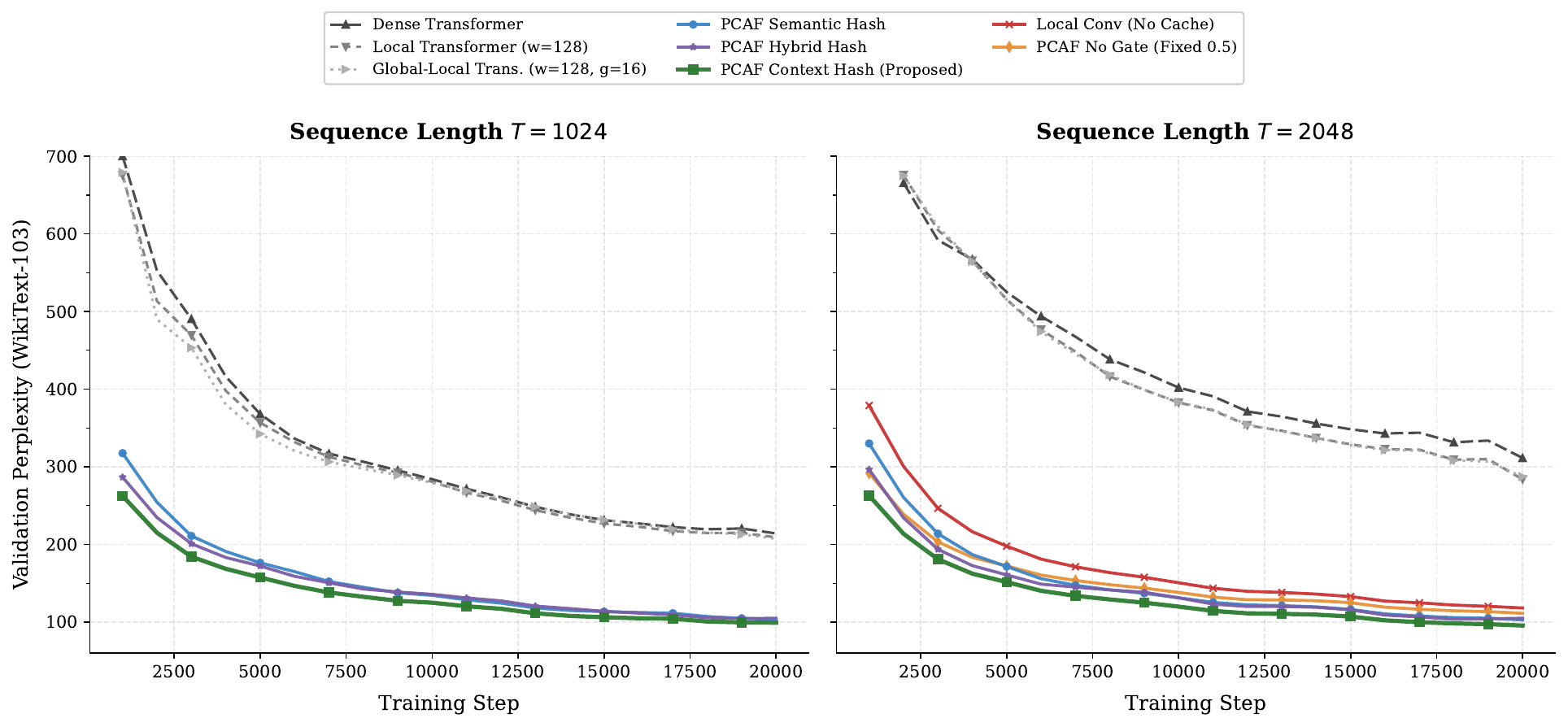}
  \caption{Next-token validation perplexity curves on WikiText-103 at sequence lengths $T = 1024$ (left) and $T = 2048$ (right) on a TPU v4-32 cluster.}
  \label{fig:ablation_curves}
  \vspace{-1.0em}
\end{figure}

\begin{figure}[H]
  \centering
  \includegraphics[width=0.50\linewidth]{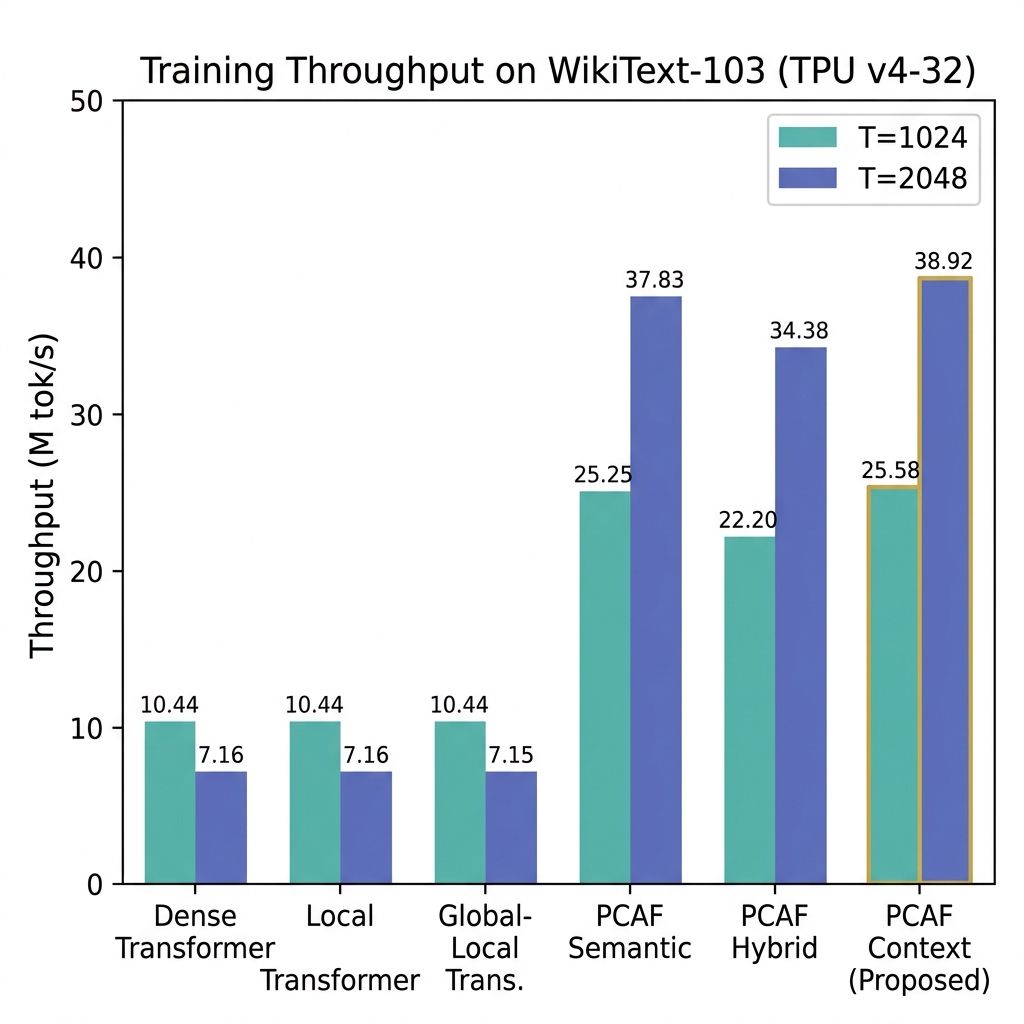}
  \caption{Training throughput comparison for the next-token diagnostic runs on WikiText-103 (TPU v4-32).}
  \label{fig:throughput}
  \vspace{-0.8em}
\end{figure}

Before the full autoregressive runs, we evaluated next-token prediction from a
single sampled context window to stress routing behavior and long-context
throughput. These diagnostics are not the headline benchmark, but they isolate
the associative-memory mechanism at high throughput. Tables~\ref{tab:tpu_baselines}
and~\ref{tab:tpu_ablations} show that PCAF-context improves over dense,
local, and global-local attention in this setting, while
Figure~\ref{fig:ablation_curves} and Figure~\ref{fig:throughput} visualize the
learning curves and throughput.

\begin{table}[!htbp]
  \centering
  \caption{Next-token diagnostic baselines on WikiText-103 using TPU v4-32. Throughput is reported in million tokens per second across the full pod.}
  \label{tab:tpu_baselines}
  \begingroup
  \tableformat
  \resizebox{\linewidth}{!}{
  \begin{tabular}{@{}llrrrrrrr@{}}
    \toprule
    Model Class \& Config & Routing/Attention Mode & Seq. Len ($T$) & Params & Final PPL & Best PPL & Eval Acc. & Throughput & Elapsed (min) \\
    \midrule
    \textbf{transformer\_dense} & Full Self-Attention & 1024 & 41.51M & 214.14 & 214.14 & 0.2280 & 10.44 M tok/s & 4.63 \\
                                &                     & 2048 & 41.71M & 311.40 & 311.40 & 0.2075 & 7.16 M tok/s & 6.63 \\
    \midrule
    \textbf{local\_transformer} & Local Attention ($w=128$) & 1024 & 41.51M & 209.08 & 209.08 & 0.2297 & 10.44 M tok/s & 4.64 \\
                                &                           & 2048 & 41.71M & 283.76 & 283.76 & 0.2155 & 7.16 M tok/s & 6.64 \\
    \midrule
    \textbf{global\_local\_trans} & Global-Local ($w=128, g=16$) & 1024 & 41.51M & 207.33 & 207.33 & 0.2310 & 10.44 M tok/s & 4.67 \\
                                  &                              & 2048 & 41.71M & 287.39 & 287.39 & 0.2103 & 7.15 M tok/s & 6.67 \\
    \midrule
    \textbf{local\_conv} (No cache) & N/A & 2048 & 41.74M & 118.12 & 118.12 & 0.2676 & \textbf{45.82 M tok/s} & \textbf{4.18} \\
    \bottomrule
  \end{tabular}
  }
  \endgroup
\end{table}

\begin{table}[!htbp]
  \centering
  \caption{Next-token PCAF routing and gating diagnostics on WikiText-103 using TPU v4-32.}
  \label{tab:tpu_ablations}
  \begingroup
  \tableformat
  \resizebox{\linewidth}{!}{
  \begin{tabular}{@{}llrrrrrrr@{}}
    \toprule
    Model Class \& Config & Routing Mode & Seq. Len ($T$) & Params & Final PPL & Best PPL & Eval Acc. & Throughput & Elapsed (min) \\
    \midrule
    \textbf{pcaf\_no\_gate} & Fixed Cache Weight ($0.5$) & 2048 & 41.74M & 111.25 & 111.25 & 0.2591 & 38.72 M tok/s & 4.95 \\
    \midrule
    \textbf{pcaf\_semantic} & Semantic Hash Routing & 1024 & 41.74M & 103.42 & 102.02 & 0.2684 & 25.25 M tok/s & 3.91 \\
                            &                       & 2048 & 41.74M & 102.84 & 102.84 & 0.2715 & 37.83 M tok/s & 5.09 \\
    \midrule
    \textbf{pcaf\_hybrid}   & Token + Semantic Mix & 1024 & 41.74M & 104.94 & 100.84 & 0.2661 & 22.20 M tok/s & 4.40 \\
                            &                      & 2048 & 41.74M & 104.90 & 101.84 & 0.2675 & 34.38 M tok/s & 5.54 \\
    \midrule
    \textbf{pcaf\_context} (Proposed) & Context Token Hash Routing & 1024 & 41.74M & \textbf{99.20} & \textbf{96.45} & \textbf{0.2768} & 25.58 M tok/s & \textbf{3.79} \\
                                      &                            & 2048 & 41.74M & \textbf{95.49} & \textbf{95.49} & \textbf{0.2804} & \textbf{38.92 M tok/s} & 4.91 \\
    \bottomrule
  \end{tabular}
  }
  \endgroup
\end{table}

\FloatBarrier

\subsection{Single-GPU Component Ablations on WikiText-2}

Tables~\ref{tab:main1024} and~\ref{tab:main2048} report controlled WikiText-2
ablations on an RTX 3060. These experiments are supporting diagnostics: they
show that removing the cache or replacing the learned gate with a fixed
mixture degrades perplexity, while retaining most of the local model's
throughput.

\begin{table}[!htbp]
  \centering
  \caption{WikiText-2 next-token prediction at context length 1024. Lower
  perplexity is better. Best results in \textbf{bold}.}
  \label{tab:main1024}
  \begingroup
  \tableformat
  \resizebox{\linewidth}{!}{
  \begin{tabular}{@{}lrrrrrrr@{}}
    \toprule
    Model & Batch & Params & Final PPL & Best PPL & Acc. & Tok/s & Peak MB \\
    \midrule
    PCAF-context         & 64 & 26.59M & \textbf{223.95} & \textbf{223.95} & \textbf{0.2156} & 442,594 & 2,410 \\
    PCAF no gate         & 64 & 26.59M & 259.53 & 259.53 & 0.2041 & 449,257 & 2,409 \\
    PCAF no cache        & 64 & 26.59M & 283.27 & 283.27 & 0.2019 & 490,490 & 2,347 \\
    Local conv only      & 64 & 26.43M & 287.86 & 287.86 & 0.1997 & \textbf{493,795} & 2,347 \\
    Local attn, $w{=}128$ & 32 & 26.71M & 357.99 & 346.95 & 0.1819 & 80,003 & 2,682 \\
    Global-local attn    & 32 & 26.71M & 364.14 & 349.66 & 0.1656 & 80,001 & 2,682 \\
    Dense Transformer    & 32 & 26.71M & 392.32 & 386.14 & 0.1625 & 112,774 & 2,678 \\
    \bottomrule
  \end{tabular}
  }
  \endgroup
\end{table}

\begin{table}[!htbp]
  \centering
  \caption{WikiText-2 next-token prediction at context length 2048.
  PCAF-context benefits from the longer context while sparse attention
  becomes slower and less accurate under this memory budget.}
  \label{tab:main2048}
  \begingroup
  \tableformat
  \resizebox{\linewidth}{!}{
  \begin{tabular}{@{}lrrrrrrr@{}}
    \toprule
    Model & Batch & Params & Final PPL & Best PPL & Acc. & Tok/s & Peak MB \\
    \midrule
    PCAF-context         & 64 & 26.59M & \textbf{210.73} & \textbf{210.73} & \textbf{0.2172} & 469,045 & 4,468 \\
    PCAF no gate         & 64 & 26.59M & 247.00 & 247.00 & 0.2062 & 472,143 & 4,467 \\
    Local conv only      & 64 & 26.43M & 260.20 & 260.20 & 0.2041 & 522,086 & 4,252 \\
    PCAF no cache        & 64 & 26.59M & 261.64 & 261.64 & 0.1975 & \textbf{524,599} & 4,252 \\
    Local attn, $w{=}128$ & 32 & 26.71M & 311.73 & 311.73 & 0.1800 & 51,010 & 4,932 \\
    Global-local attn    & 32 & 26.71M & 316.45 & 316.45 & 0.1775 & 50,827 & 4,932 \\
    Dense Transformer    &  8 & 26.71M & 577.42 & 458.20 & 0.1225 & 77,095 & 2,713 \\
    \bottomrule
  \end{tabular}
  }
  \endgroup
\end{table}
\FloatBarrier

\section{Discussion}

The results support three broader observations. First, separating memory
addressing from value resolution enables a favorable speed--quality trade-off.
At 303M parameters, PCAF-semantic achieves the best perplexity on both
WikiText-103 and PG-19 while running about 1.4--2.0$\times$ faster than dense
and local attention baselines on the TPU v4-32 pod. The smaller diagnostic runs
show a larger raw throughput gap because the associative retrieval path scales
more favorably with context length than dense all-pairs attention.

Second, the combination of a parametric local path and a non-parametric cache
is more effective than either component alone. The local convolutional path
provides smooth syntactic compositionality for common $n$-gram contexts, while
the cache provides direct long-range token lookup for rarer patterns. The
learned gate adaptively weights these two sources, a behavior consistent with
prior work showing that cache language models benefit most at positions where
the parametric model is uncertain
\citep{grave2016continuous,khandelwal2019knn}.

Third, routing behavior changes with scale. At 41M parameters, the discrete
context route is highly competitive and often strongest in diagnostic
next-token settings. At 303M parameters under full autoregressive training,
the learned semantic route gives the best perplexity. This suggests that
continuous routing benefits from larger hidden representations, while discrete
context hashing remains a fast and reliable retrieval path.

\section{Limitations and Future Work}
\label{sec:limitations}

Several limitations of the current work merit discussion. First, while we have
successfully scaled our evaluation to large-scale, full-sequence autoregressive
pretraining at 303M parameters (Section~\ref{sec:full-ar-303m}), evaluating
downstream zero-shot generation tasks, larger text corpora such as the Pile,
and alternative sequence modalities such as audio or code would further
strengthen the claims.

Second, the current address is a polynomial $n$-gram hash, which provides
effective candidate selection but cannot generalize to semantically similar
contexts with different surface forms. Future work should explore learned
address functions, such as a small encoder projecting contexts into discrete
codebook entries, while retaining the sparse memory primitive.

Third, the Mamba baseline could not be evaluated at batch size 64 due to GPU
memory constraints on this hardware. A complete efficiency comparison requires
smaller-batch Mamba runs and normalization by tokens processed, wall-clock
time, and peak memory simultaneously.

Fourth, the hash-bucket scheme discards records beyond the per-bucket capacity
$K$. Under adversarial or highly repetitive inputs, many records could collide
into a single bucket, reducing effective recall. An analysis of bucket
occupancy statistics and the impact of bucket capacity on recall would
clarify the robustness of the addressing scheme.

\section{Conclusion}

This paper introduces PCAF, a gated sparse associative-memory language
model that replaces dense token-to-token attention with a combination of a
local causal convolutional path and a hash-bucket successor-token cache. The
model is trained end-to-end with a single cross-entropy loss, with no
pre-built external index. On controlled WikiText-2 next-token prediction experiments,
PCAF gives lower perplexity than dense attention, sliding-window attention,
global-local attention, and local-convolution ablations at both 1024- and
2048-token contexts, while also providing higher training throughput than the
attention baselines. On the larger WikiText-103 and PG-19
corpora under a full autoregressive pretraining objective on a distributed TPU v4-32 cluster,
our 303M-parameter PCAF-semantic model reaches validation perplexities of
\textbf{33.30 PPL} at $T=1024$ and \textbf{36.31 PPL} at $T=2048$ on WikiText-103 (outperforming the
dense Transformer baseline by up to \textbf{16.22 points}), and \textbf{50.94 PPL} at $T=1024$ and \textbf{52.45 PPL} at $T=2048$ on PG-19.
Simultaneously, PCAF-semantic delivers a \textbf{\(1.42\times\) to \(2.02\times\) higher throughput}
(0.89M and 0.61M tokens/second vs. 0.44M and 0.43M tokens/second on WikiText-103) than matched attention baselines.
Ablation studies show that both the associative cache and the learned mixture
gate are necessary contributors to these gains. These results suggest that
parallel content-addressed memory warrants further study as a primitive for
efficient long-context sequence modeling.

\bibliographystyle{plainnat}
\bibliography{references}

\end{document}